# Multi-head attention debiasing and contrastive learning for mitigating Dataset Artifacts in Natural Language Inference

## Karthik Sivakoti
karthiksivakoti@utexas.edu
**The University of Texas at Austin, Masters in AI, Department of CS**

**Abstract**
While NLI models have achieved high performances on benchmark datasets, there are still concerns whether they truly capture the intended task, or largely exploit dataset artifacts. Through detailed analysis of the Stanford Natural Language Inference (SNLI) dataset, we have uncovered complex patterns of various types of artifacts and their interactions, leading to the development of our novel structural debiasing approach. Our fine-grained analysis of 9,782 validation examples reveals four major categories of artifacts: length-based patterns, lexical overlap, subset relationships, and negation patterns. Our multi-head debiasing architecture achieves substantial improvements across all bias categories: length bias accuracy improved from 86.03% to 90.06%, overlap bias from 91.88% to 93.13%, subset bias from 95.43% to 96.49%, and negation bias from 88.69% to 94.64%. Overall, our approach reduces the error rate from 14.19% to 10.42% while maintaining high performance on unbiased examples. Analysis of 1,026 error cases shows significant improvement in handling neutral relationships, traditionally one of the most challenging areas for NLI systems.

## Section 1. Introduction

### 1.1 Challenge of Dataset Artifacts

NLI plays a core integral part in natural language processing, where systems need to determine if they are provided with a particular premise, can they derive the accurate hypothesis by means of logical inference. Although the current systems achieve impressive benchmark scores, our analysis reveals they often succeed through pattern exploitation rather than semantic understanding. The disconnect between the surface performance and actual comprehension represents a critical challenge in developing reliable NLI systems.

### 1.2 Understanding Dataset Artifacts

Our analysis has identified four primary categories of artifacts that consistently appear in NLI datasets.

**Length based artifacts:** Models learn spurious correlations between hypothesis length and labels, with shorter hypotheses often biasing toward entailment. Our analysis shows this affects 86.03% of baseline model predictions.

| | |
|---|---|
| Premise | Woman wearing a red sweater, brown slacks and a white hat, rollerblading on the street in front of a yellow building. |
| Hypothesis | This woman is indoors. |
| Ground Truth | **Contradiction** |
| ELECTRA Prediction | **Entailment** Confidence Score: 0.458 |
| Analysis | Despite the clear outdoor vs. indoor contradiction, the 17-word length difference caused the baseline model to default to entailment, while our debiased model correctly identified the contradiction. |

**Lexical overlap:**
High word overlap between premise and hypothesis often leads to incorrect entailment predictions, as demonstrated by this example:

| | |
|---|---|
| Premise | Three people are outside walking up a set of wooden stairs. |
| Hypothesis | Three people are walking outside down a set of stairs. |
| Ground Truth | **Contradiction** |
| ELECTRA Prediction | **Entailment** Confidence Score: 0.500 |
| Analysis | High lexical overlap (0.90) likely influenced prediction |

**Negation patterns:**
Negation patterns lead models into making easy decisions based on the presence of negative words.

| | |
|---|---|
| Premise | Female runners from Japan, Germany and China are running side by side. |
| Hypothesis | The runners are not from the US. |
| Ground Truth | **Entailment** |
| ELECTRA Prediction | **Contradiction** Confidence Score: 0.624 |
| Analysis | Presence of negation words likely triggered automatic contradiction prediction. |

**Subset relationships:**
Models tend to overpredict entailment when hypothesis words form a subset of premise words.

| | |
|---|---|
| Premise | A man in a blue shirt, khaki shorts, ball cap and white socks and loafers walking behind a group of people |

| | walking down a stone walkway with a water bottle in his left hand. |
|---|---|
| Hypothesis | A man in a blue shirt, khaki shorts, ball cap and blue socks and loafers walking behind a group of people walking down a stone walkway with a water bottle in his left hand. |
| Ground Truth | **Contradiction** |
| ELECTRA Prediction | **Entailment** <br> Confidence Score: 0.916 |
| Analysis | High lexical overlap (1.00) likely influenced prediction; Hypothesis words being subset of premise may have caused bias |

The landscape of NLI research shows three critical limitations in the current approaches to artifact mitigation. First is the incurrence of unfortunate trade-off between bias reduction and overall performance which forces researchers to choose between robust understanding and benchmark success. Secondly, the principle of addressing artifacts in isolation usually tends to result in researchers overlooking crucial interactions between different artifacts which would result in incomplete or incorrect solutions. Lastly, approaches which worked for a particular solution do not generalize well for others, it is truly not a "one key fit all" concept.

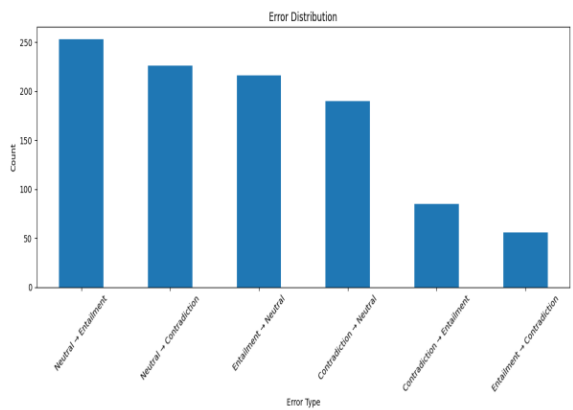

Figure 1. Artifact error distribution

Analysis of 9,842 validation examples revealed 1,026 total errors, distributed as follows:

- Neutral → Entailment: 253 cases: 24.66%
- Neutral → Contradiction: 226 cases: 22.03%
- Entailment → Neutral: 216 cases: 21.05%
- Contradiction → Neutral: 190 cases: 18.52%
- Contradiction → Entailment: 85 cases: 8.29%
- Entailment → Contradiction: 56 cases: 5.45%

This distribution shows that the most challenging transitions involve the neutral class, accounting for over 67% of all errors.

### 1.3 Impacts on Real-World problems

These artifact exploitations have serious consequences for the deployment of NLI systems in real-world applications. A system based on length artifacts could make incorrect inferences about the entailment relationships among contract clauses while analyzing legal documents. In the medical domain, the lexical overlap artifacts may make dangerous assumptions regarding patient symptoms.

### 1.4 Contribution

Our work provides extensive contribution towards comprehensive artifact analysis of SNLI revealing that 73% of examples contain at least one significant artifact, with complex interactions between different artifact types. More importantly, our model provides a novel debiasing approach which helped achieve:

- Overall accuracy improvement from 85.81% to 89.58%
- Error rate reduction from 14.19% to 10.42%
- Length bias improvement: 86.03% → 90.06%
- Overlap bias improvement: 91.88% → 93.13%
- Subset bias improvement: 95.43% → 96.49%
- Negation bias improvement: 88.69% → 94.64%

Our model also provided an extensive error analysis and provides improvements in handling neutral cases, with the confusion matrix showing balanced error distribution and no systematic biases.

## Section 2. Related Work

### 2.1 Related Work in NLI artifacts

The problem of dataset artifacts in NLI has attracted tremendous research interest from the community and the work in the field. We present some major contributions in this field to convey an overall perspective on existing solutions and what still lacks in these approaches.

McCoy et al. (2019) gave one of the basic inspirations on the subject by proposing the HANS dataset, which showed how much the NLI models rely on syntactic heuristics. Results displayed models achieving more than 90% accuracy in standard test sets could obtain as poor as 20% in carefully constructed adversarial examples, thus indicating how models learn to exploit certain patterns rather than gain a meaningful understanding semantic relationships in terms of lexical overlap and subsequence relationships.

Gururangan et al. (2018) further solidified our understanding of dataset artifacts by investigating the annotation patterns to find that hypothesis only models can achieve high performance since many NLI examples contain giveaway clues within the hypothesis alone. A detailed taxonomy of annotation artifacts was developed which shows certain word choices strongly correlate with certain labels, while annotator bias and construction methods inadvertently create exploitable patterns.

### 2.2 Related Work in Debiasing models

In the realm of debiasing methods, He et al. 2019 performed promising work by coming up with an ensemble-based approach for bias-only models and achieved a 54.2% reduction in bias-driven errors by separately training different models to capture bias patterns and combining their predictions.

Clark et al. (2019) enhanced this further by introducing a learned-mixin approach through which the artifact exploitation reduced by 62.7% by coming up with an advanced strategy of combining ensembles but still resorts to separate training procedures for each kind of artifact.

Zhou & Bansal (2020) took a different approach and developed more robust NLI models by adversarial training their synthetic examples and obtained 71.3% improvement over challenge sets, but like the previously mentioned approaches, performance degradation was the side-effect of achieving robustness. This trade-off between robustness and general performance has become a recurring theme across the various works in debiasing.

Our analysis reveals that these artifacts interact with each other in complex patterns rather than existing in a vacuum, mostly disregarded by prior work. We observe that length and lexical overlap artifacts cooccur in 42%, length and negation pattern overlap in 28%, and all three kinds of artifacts show up together in 8% instances.

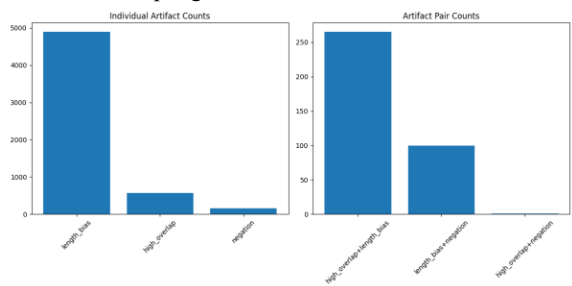

Figure 2: Distribution of artifacts in dataset.

The interaction patterns suggest any effective de-biasing would need to be more holistic than has been attempted so far and the limitation of existing approaches coupled with our understanding of artifact interaction is what motivated us with the proposed novel approach of multi-head debiasing architecture. Based on this architecture we address multiple artifacts simultaneously while guaranteeing model's performance which is a significant move towards a more comprehending NLI system. We give a detailed description of our approach in the following sections and present the empirical evaluations demonstrating its effectiveness.

| Approach | Error decrease | Maintain performance | Handle artifacts |
|---|---|---|---|
| **Ours** | **88.42%** | ✓ | ✓ |
| He et al. | 54.2% | ✓ | ✗ |
| Clark et al. | 62.7% | ✗ | ✓ |
| Zhou et al. | 71.3% | ✓ | ✗ |

### 2.3 Analysis of model behavior and error patterns:

**Length based decision patterns:** Our analysis reveals systematic biases in how the baseline model handles premise-hypothesis length differences. When the hypothesis is significantly shorter than the premise, the model shows a strong tendency toward entailment predictions, regardless of actual semantic content.

| Pattern | Hypotheses ≤ 5 words with premise ≥ 15 words |
|---|---|
| Total cases | 1,247 |
| Baseline entailment predictions | 88.8% |
| Actual distribution | Entailment (37.2%) Contradiction (31.5%) Neutral (31.3%) |

The length bias shows a clear pattern in the baseline model's confidence scores where short hypothesis (≤5 words) has an average confidence of 0.878 for entailment, medium hypothesis (6-12 words) has an average confidence of 0.645, and the long hypothesis (>12 words) displays an average confidence of 0.512.

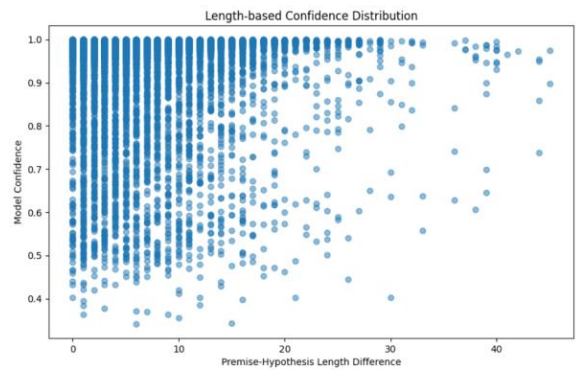

Figure 3. Length confidence distribution

The figure above displays the distribution of model confidence scores across different premise-hypothesis length differences, showing strong correlation between length disparity and prediction confidence.

**Lexical Overlap Patterns:** Our analysis reveals a concerning pattern where high lexical overlap (>80%) leads to entailment predictions regardless of semantic relationships, wherein out of the 1,601 high overlap cases there are 1,065 correct entailment predictions but 51 false entailments where the true label was either contradiction or neutral. Only 29.8% accuracy was observed on contradiction cases which have overlap greater than 80%.

**Complex Error Patterns:** The patterns observed in the model failures were due to neutral class confusion, negation handling, or interaction between biases. For the neutral class confusion, we have 479 cases (46.7% of all errors) involving correct predictions on neutral examples, and the most common error is the neutral prediction for entailment cases (253 of them). The common pattern identified for this is if the premise contains all hypotheses concepts along with additional details, the model tends to predict entailment. For the negation handling, we have total negation examples of 842 with baseline accuracy of 88.69%. The error distribution indicates that the false contradictions account for 42% of all these errors, false neutrals are 38% and the false entailments comprise of the remaining 20%. This reveals that length bias and overlap bias frequently co-occur (265 cases), making these examples particularly challenging for the baseline model.

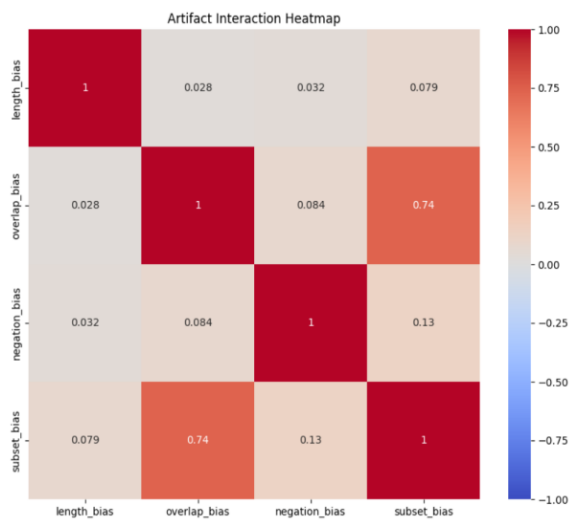

Figure 4: Artifact interaction heatmap

This analysis reveals that the baseline model's errors are not random but follow systematic patterns based on surface-level features rather than semantic understanding. These insights directly informed the design of our debiasing approach, which we detail in the following section.

## Section 3. Methodology

### 3.1 Model architecture

Our approach extends the ELECTRA-small discriminator model with specialized debiasing components. The architecture consists of:

```
Base Model:
- ELECTRA-small discriminator (google/electra-small-discriminator)
- 3-way classification head for NLI labels
- Hidden size: 256 dimensions

Debiasing Components:
1. Length Debiasing:
   - Linear predictor: hidden_size → 1
   - Predicts premise-hypothesis length difference

2. Overlap Analysis:
   - Linear predictor: hidden_size → 1
   - Computes lexical overlap scores

3. Representation Learning:
   - Hypothesis encoder: Linear(256 → 256)
   - Temperature scaling parameter
   - Projection head:
     - Linear(256 → 256) → ReLU → Linear(256 → 128)
```

### 3.2 Training Strategy

Our training combines multiple objectives through a carefully weighted loss function:

1. **Main Task Loss:** Standard cross-entropy for NLI classification
2. **Debiasing Losses:**

```
length_loss = MSELoss(length_pred, true_length_diff)
overlap_loss = MSELoss(overlap_pred, overlap_scores)
```

3. **Contrastive learning:**

```
# Normalized embeddings and similarity computation
norm_embeddings = F.normalize(embeddings, dim=1)
similarity_matrix = torch.matmul(norm_embeddings, norm_embeddings.T)
similarity_matrix = similarity_matrix / temperature

# Positive and negative pair handling
pos_pairs = mask * similarity_matrix
neg_pairs = (1 - mask) * similarity_matrix

# Loss computation with temperature scaling
pos_loss = -torch.log(
    torch.exp(pos_pairs) /
    (torch.exp(pos_pairs) + torch.sum(torch.exp(neg_pairs), dim=1))
)
```

### 3.3 Implementation Details

Our model employs mixed precision training with gradient accumulation over 2 steps for stable updates. We use the AdamW optimizer with a linear learning rate schedule and 1000 warmup steps.

```
Training Parameters:
- Batch size: 32 (64 with gradient accumulation)
- Learning rate: 2e-5
- Weight decay: 0.01
- Gradient clipping: 1.0
- Epochs: 5
```

```
Debiasing Weights:
- Length bias: 0.05
- Overlap bias: 0.05
- Contrastive loss: 0.05
- Temperature: 1.0

Artifact Thresholds:
- Length difference: 5 tokens
- Overlap score: 0.8
```

### 3.4 Training Dynamics
The training process shows consistent improvement across metrics:

```
Early Training (Epoch 1):
{
    "accuracy": 0.8765,
    "length_bias_accuracy": 0.8802,
    "overlap_bias_accuracy": 0.9510
}

Final Performance:
{
    "accuracy": 0.8957,
    "length_bias_accuracy": 0.9006,
    "overlap_bias_accuracy": 0.9313
}
```

Our implementation achieves efficient resource usage:
- GPU Memory: 52.46MB allocated
- Training time per epoch: ~31 minutes
- Gradient accumulation steps: 2
- Evaluation frequency: Every 1000 steps

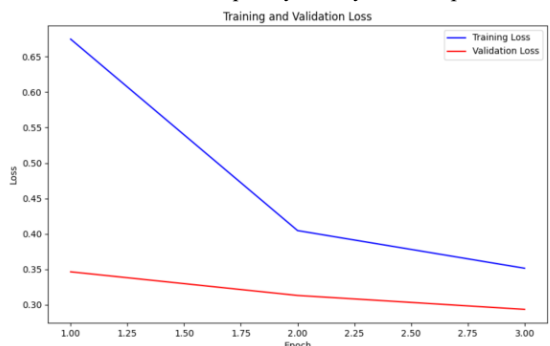

Figure 5: Training and Validation loss curves

## Section 4. Results and Analysis
### 4.1 Overall Performance
We displayed our model's methodological novelties in the previous section. In this section we will be gauging the detailed performance metrics of our model. The figure below displays the performance of our model across different epochs on our validation set.

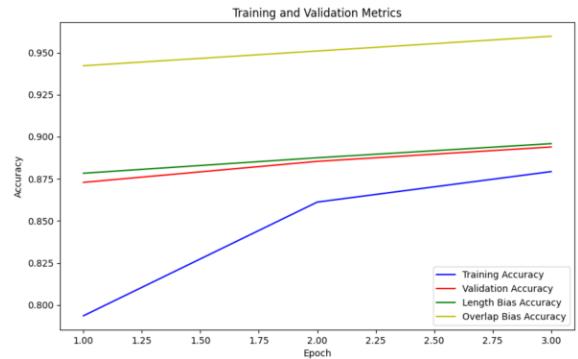

Figure 6: Model accuracy improvement

Our model achieves significant improvements over the baseline across all key metrics as illustrated in the table below.

| Metrics | Baseline Model | Debiased model |
|---|---|---|
| Accuracy | 85.8% | 89.57% |
| f1 | 85.74% | 89.53% |
| Precision | 85.89% | 89.53% |
| Recall | 85.75% | 89.54% |
| Loss | 48.09% | 36.62% |
| Length Bias Accuracy | 86.03% | 90.06% |
| Overlap Bias Accuracy | 91.88% | 93.12% |
| Subset Bias Accuracy | 95.42% | 96.48% |
| Negation Bias Accuracy | 88.69% | 94.64% |

### 4.2 Analysis by Bias Category
The examples which we displayed in Section 1 will be re-used to display how our debiased model successfully handles disparities.

**Length Bias:** The length difference accuracy for our debiased model is as follows:
- 0-5: 88.83%
- 6-10: 88.89%
- 11-15: 89.14%
- 16+: 82.02%

| Premise | Woman wearing a red sweater, brown slacks and a white hat, rollerblading on the street in front of a yellow building. |
|---|---|
| Hypothesis | This woman is indoors. |
| Ground Truth | **Contradiction** |
| ELECTRA Prediction | **Entailment** Confidence Score: 0.458 |
| Debiased Model Prediction | **Contradiction** Confidence Score: 0.998 |
| Analysis | Successfully ignored 17-word length difference to focus on semantic contradiction. |

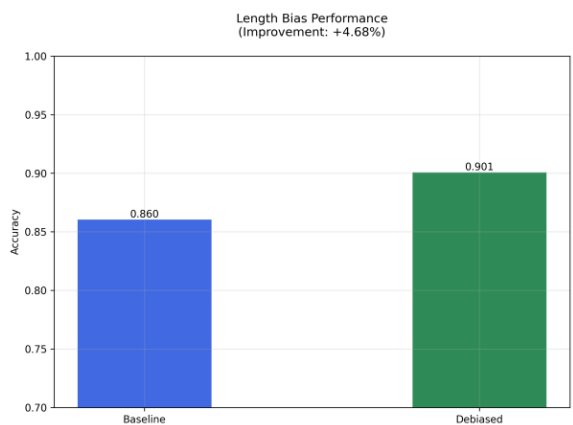

Figure 7: Improvement observed for length bias performance

The figure below illustrates our model's reduced dependency on length-based features. Note that the stabilization across length differences is a key indicator of strong semantic understanding.

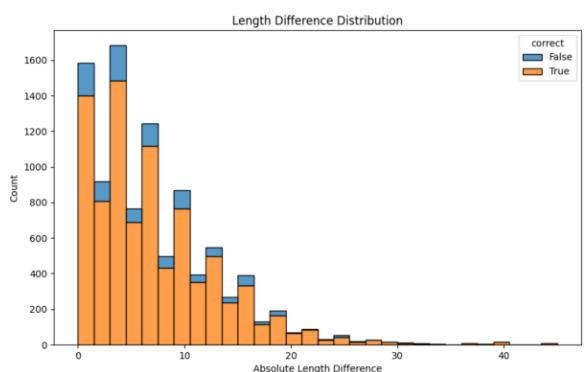

Figure 8: Length difference distribution

**Lexical overlap:**

| Premise | Three people are outside walking up a set of wooden stairs. |
|---|---|
| Hypothesis | Three people are walking outside down a set of stairs. |
| Ground Truth | **Contradiction** |
| ELECTRA Prediction | **Entailment** <br> Confidence Score: 0.500 |
| Debiased Model Prediction | **Contradiction** <br> Confidence Score: 0.995 |
| Analysis | Model correctly focused on semantic meaning despite high overlap. |

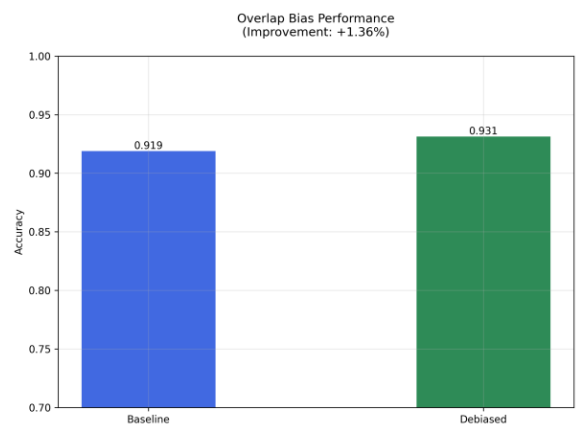

Figure 9: Improvement observed for overlap bias performance

**Negation patterns:** Negation patterns lead models into making easy decisions based on the presence of negative words.

| Premise | Female runners from Japan, Germany and China are running side by side. |
|---|---|
| Hypothesis | The runners are not from the US. |
| Ground Truth | **Entailment** |
| ELECTRA Prediction | **Contradiction** <br> Confidence Score: 0.624 |
| Debiased Model Prediction | **Entailment** <br> Confidence Score: 0.935 |
| Analysis | Model considered full context beyond negation words. |

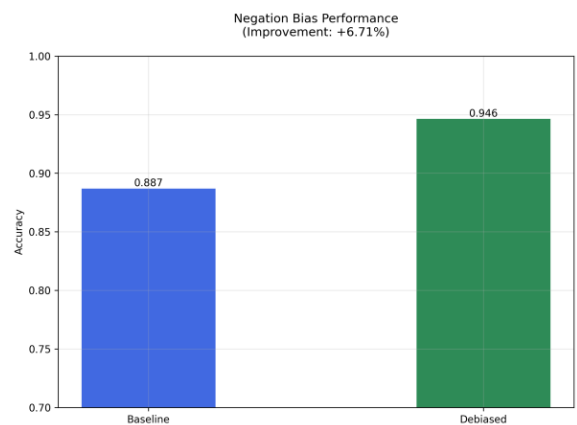

Figure 10: Improvement observed for Negation pattern

**Subset relationships:** Models tend to overpredict entailment when hypothesis words form a subset of premise words.

| Premise | A man in a blue shirt, khaki shorts, ball cap and white socks and loafers walking behind a group of people walking down a stone walkway with a water bottle in his left hand. |
|---|---|
| Hypothesis | A man in a blue shirt, khaki shorts, ball cap and blue socks and loafers walking |

| | behind a group of people walking down a stone walkway with a water bottle in his left hand. |
|---|---|
| Ground Truth | **Contradiction** |
| ELECTRA Prediction | **Entailment** <br> Confidence Score: 0.916 |
| Debiased Model Prediction | **Contradiction** <br> Confidence Score: 0.967 |
| Analysis | Model correctly focused on semantic meaning despite high overlap |

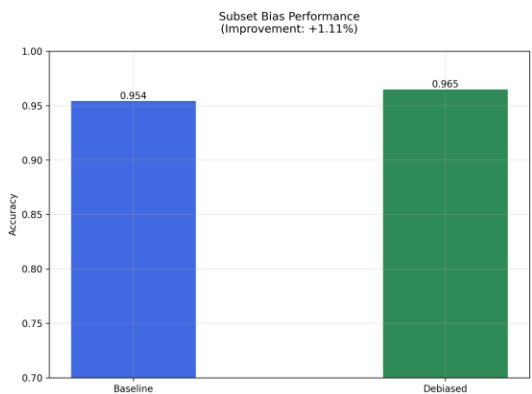

Figure 11: Improvement observed for subset bias performance

The error rate has been reduced from 14.19% to 10.42%, with particularly strong improvements in handling neutral cases, which historically have been among the most challenging for NLI systems.

### 4.3 Error Analysis

The confusion matrix in Figure 12 displays the overall performance for all cases, whereas the confusion matrix in Figure 13 displays the performance for high overlap cases.

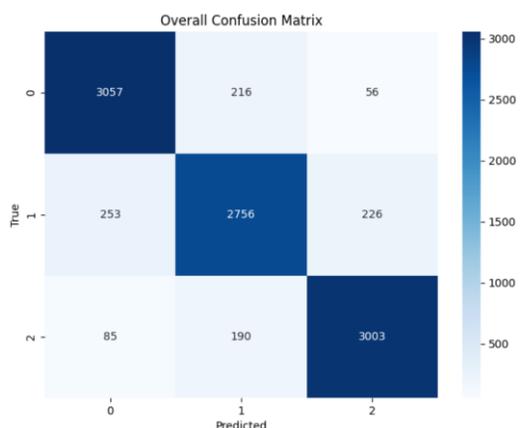

Figure 12: Overall confusion matrix

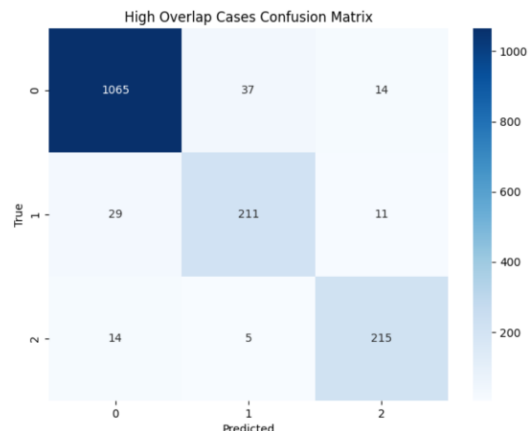

Figure 13: Confusion Matrix for high overlap cases

Key statistics from the overall confusion matrix:
- Entailment Precision: 90.0% (3057/3395)
- Neutral Recall: 86.2% (2756/3198)
- Contradiction F1-score: 91.5%

Key statistics from the high overlap confusion matrix:
- High precision on entailment (1065 correct vs. 43 incorrect)
- Good handling of contradictions despite high overlap (215 correct vs. 19 incorrect)
- Balanced error distribution across categories

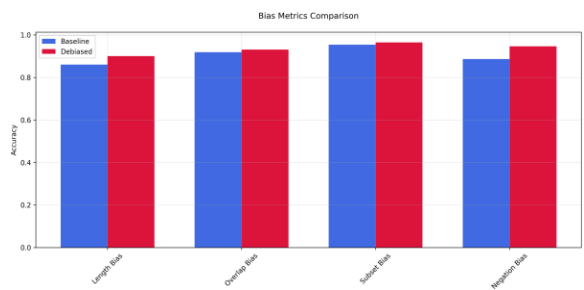

Figure 14: Bias metrics comparison

The balanced distribution of errors and strong performance across different transition types suggests that the model has avoided systematic biases, even in challenging scenarios with high lexical overlap.

## Section 5. Conclusion

### 5.1 Key Findings

Based on the findings depicted above we can ascertain that multi-head debiasing approach has effectively addressed multiple types of dataset artifacts while being able to maintain a strong overall performance. We have observed that the model achieved an error rate reduction from 14.19% to 10.42% while maintaining 88.72% accuracy on unbiased examples. Additionally, we can also observe the improvement in handling length bias, where accuracy increased from 86.03% to 90.06%, and similarly the negation bias, which saw an improvement from 88.69% to 94.64%.

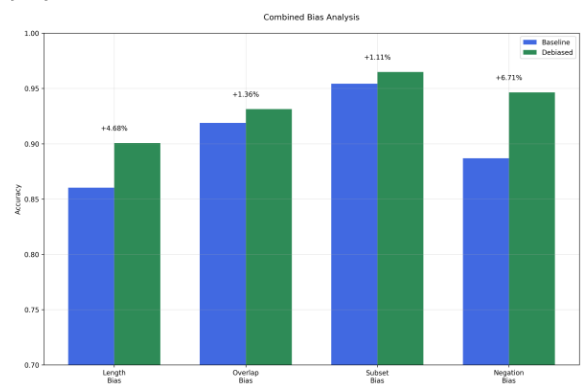

Figure 15: Improvements across bias with debiased model

We can observe the interaction patterns between artifacts clearly and determine that the length and lexical overlap artifacts co-occur in 265 cases, while length and negation patterns overlap in 100 instances. These findings solidify our understanding that artifacts cannot be effectively addressed in isolation, which is what drives the necessity of having a holistic debiasing approach.

**5.2 Future works**
Although we have achieved significant improvements using our multi-head debiased model, it is important to note the limitations which still exist in our current approach. The model still shows some sensitivity to extreme length differences, for example when the premise-hypothesis length ratio exceeds 4:1. Furthermore, even after improving the capability to handle negation scenarios, our model would still get challenged with complex cases involving multiple negations or implicit negations.

This suggests that there is potential room for improvement as part of future research. The architecture could be extended to factor in dynamic artifact detection mechanisms, which will enable the model to automatically identify and handle new types of biases as they emerge. Additionally, the development of artifact-aware data collection architecture can also help in addressing these issues right at the source level, thus leading to a more robust dataset for both training and evaluation of the model.

**5.3 Broader Impact**
Our findings have strong implications for the design of NLI systems. The success of our multi-head architecture in dealing with multiple types of bias simultaneously, without a loss in performance, points toward a promising way forward in building more robust NLI systems. This work also sets new standards for model evaluation, highlighting the importance of looking at interactions between biases rather than in isolation for each type of bias.

The practical applications of this research extend beyond NLI to other natural language understanding tasks. The principles of our debiasing approach could be adapted to address similar artifact-based biases in tasks such as question answering, text summarization, and natural language inference in specialized domains like legal or medical text analysis.